\documentclass[letterpaper]{article} 
\usepackage[preprint]{BCM}
\usepackage{times}
\usepackage[hyphens]{url}  
\usepackage{graphicx} 
\usepackage{natbib}  
\usepackage{caption} 
\usepackage{algorithm}
\usepackage{algorithmic}
\usepackage{pifont}
\usepackage{amsmath}   
\usepackage{amssymb}   
\usepackage{multirow}   
\usepackage{newfloat}
\usepackage{listings}
\DeclareCaptionStyle{ruled}{labelfont=normalfont,labelsep=colon,strut=off} 
\floatstyle{ruled}
\newfloat{listing}{tb}{lst}{}
\floatname{listing}{Listing}

\usepackage{booktabs}

\title{Compact Bellman-Grounded Cognitive Maps for Cost-Aware Navigation}
\author{
    Yuzhe Han\textsuperscript{\rm 1},
    Mingkun Xu\textsuperscript{\rm 1}\corresponding,
    Yujie Wu\textsuperscript{\rm 2}\corresponding
}
\affiliations{
    \textsuperscript{\rm 1}Guangdong Institute of Intelligence Science and Technology (GDIIST), Zhuhai, China\\
    \textsuperscript{\rm 2}Department of Computing, The Hong Kong Polytechnic University, Hong Kong SAR, China\\
    xumingkun@gdiist.cn,
    yu-jie.wu@polyu.edu.hk
}

\begin{document}

\maketitle

\begin{abstract}
Biological agents navigate familiar environments not by re-solving routes for each new goal, but by reusing a learned map built once and read off as goals change. Existing artificial cognitive-map models mimic this reuse, yet their guidance is not explicitly grounded in additive heterogeneous route costs.  Furthermore, they often struggle with memory efficiency:  representative state-indexed and high-rank spectral constructions incur substantial storage growth as the environment scales. We present BCM, which grounds a reusable cognitive map in local edge costs through a self-supervised Bellman-grounded objective and a compact coordinate encoding, supporting changing goal queries without per-goal retraining. On weighted grids of up to  $N=1600$ nodes, BCM maintains full success and only a 5\% mean Gap relative to exact Dijkstra search, compared with about $45\%$ for a connectivity-based spectral baseline.   Notably, as the graph size increases from  $N=400$ to  $N=3600$, its memory footprint grows sublinearly while maintaining competitive performance, making our method scalable to complex environments. Together, these results show that additive route costs can be written into a compact, reusable cognitive-map representation, bridging the gap between biological flexibility and optimal path planning.
\end{abstract}

\section{Introduction}
\begin{figure}[!t]
\centering
\includegraphics[width=\columnwidth]{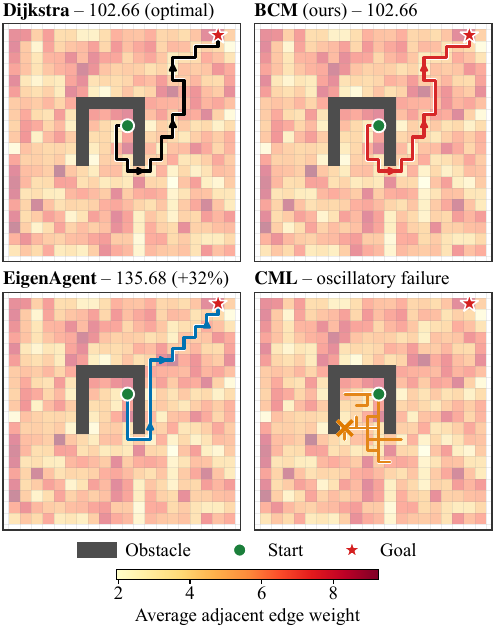}
\caption{
Planning on the same weighted U-shaped graph. BCM matches the Dijkstra
reference cost, EigenAgent incurs $32\%$ higher cost, and CML fails by
oscillation.
}
\label{fig:paths}
\end{figure}
Animals and humans navigate familiar environments not by solving each new goal from scratch, but by reusing a learned cognitive map~\cite{tolman1948cognitive}, built through experience and queried as goals change. This suggests that the difficulty of planning lies not only in search but also in the efficiency of the representation being reused. Reproducing this flexibility in artificial agents remains challenging, particularly on weighted graphs, where the most direct geometric route need not be the least costly.

Cognitive-map models were designed to provide this exact type of reusable guidance,  yet they are currently missing a crucial dimension: the actual cost of movement. For instance, successor representations capture policy-conditioned predictive occupancy~\cite{dayan1993sr,stachenfeld2017sr}; spectral methods organize states according to graph connectivity~\cite{mahadevan2007pvf,zuo2026eigenagent}; and inverse-model metrics induce action-consistent directional guidance~\cite{stoeckl2024cml,lin2026gcml}. None of these approaches inherently understands that traversing a swamp costs more than walking on a paved road. Consequently, under heterogeneous edge weights, such representations can favor routes that are geometrically or topologically plausible yet substantially more expensive (Figure~\ref{fig:paths}).

Beyond this lack of cost awareness, existing reusable maps also face scalability challenges. Full-rank spectral maps retain a quadratic number of coefficients, while CML-style learners rely on one-hot state and action parameterizations that grow with the enumerated spaces. The core challenge is therefore to learn a compact, reusable cognitive-map representation that captures additive route-cost structure.

To address this challenge, we introduce the \textbf{Bellman-grounded Cognitive Map (BCM)} that combines cost-aware local scoring with a compact, reusable, goal-independent representation for planning across goals. Three key innovations enable this: first, a Bellman-grounded objective makes the scores cost-aware: for a nonterminal step, its score is shaped by the cost of that edge plus the best downstream score. We construct this self-supervised objective from the graph's own transitions and edge costs, without shortest-path supervision. Second, BCM retains online planning: each action is selected on demand from local scores, without first constructing a complete path or expanding a search tree~\cite{mattar2022planning,russell2020aima}. And because the encoder is goal-independent, the underlying node representation is shared across goals. Changing the goal, therefore, requires neither rebuilding nor retraining the map. Third, an interaction-based coordinate encoder keeps the map compact by binding per-axis embeddings, so that the coordinate-dependent parameter count grows as $\mathcal{O}(\sqrt{N})$ on two-dimensional grids.

Our contributions are threefold:
\begin{itemize}
\item \textbf{A cost-grounded cognitive map.} We identify additive route cost as a missing representational component of reusable cognitive maps on heterogeneous weighted graphs, and introduce BCM to ground goal-conditioned guidance in weighted route structure without shortest-path supervision.
\item \textbf{Compact online multi-goal planning.} BCM supports online planning across changing goals from a shared cognitive map, while its coordinate-dependent parameter count grows as $\mathcal{O}(\sqrt{N})$ on two-dimensional grids.
\item \textbf{Empirical validation.} Across three weighted-grid layouts through $N=1600$, BCM maintains full success and approximately $5\%$ mean Gap under the widest weight range, compared with about $45\%$ for the evaluated connectivity-based spectral map.
\end{itemize}
\section{Method}
\begin{figure*}[t]
    \centering
    \includegraphics[width=1.0\textwidth]{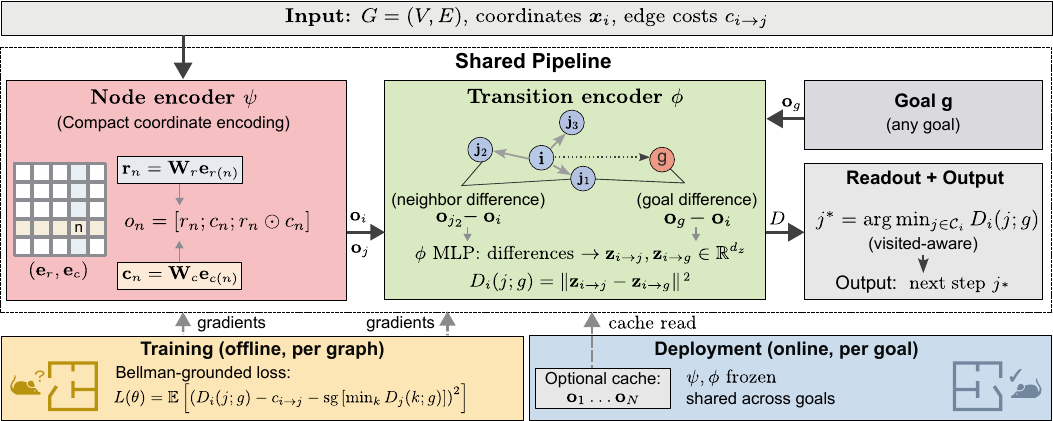}
    \caption{Overview of BCM. A shared pipeline compactly encodes node coordinates ($\psi$) and transitions from embedding differences ($\phi$). Training applies a Bellman-grounded cost-grounding objective to the discrepancy $D$. At deployment, BCM performs online local readout, while node embeddings may optionally be cached and reused across goal queries.}
    \label{fig:overall_model}
\end{figure*}

As illustrated in Figure~\ref{fig:overall_model}, BCM has three components: transition encoding, Bellman-grounded training, and online greedy readout. A node encoder $\psi$ maps node coordinates to embeddings, and a transition encoder $\phi$ maps differences between node embeddings to transition embeddings.

\subsection{Problem Setup}
We consider repeated planning in a single fixed environment, modeled as an undirected weighted graph $G=(V,E)$ with $N=|V|$ nodes. Each node $i\in V$ has coordinates $x_i$ (e.g., its position on a 2D grid), and each edge $(i,j)\in E$ has a known positive cost $c_{i\to j}>0$. We write $\mathcal{N}(i)=\{j\mid(i,j)\in E\}$ for the neighbors of $i$. Given a start node $s$ and a goal node $g$, the planner constructs a route by sequentially selecting a neighboring node until reaching $g$.

\subsection{Transition Encoding}
\label{sec:transition_encoding}

This difference-based form follows inverse-model cognitive maps~\cite{stoeckl2024cml}, but rather than aligning each difference with a separate action embedding, we compare a candidate local transition with the displacement toward the goal. A node encoder $\psi$ maps each node's coordinates to an embedding $o_i=\psi(x_i)\in\mathbb{R}^{d_o}$. The transition encoder $\phi$ then maps an embedding displacement between two nodes to a transition embedding:
\begin{equation}
z_{i\to j}=\phi(o_j-o_i),
\end{equation}
where $z_{i\to j}\in\mathbb{R}^{d_z}$.

For a candidate neighbor $j\in\mathcal{N}(i)$ and goal $g$, we define
\begin{equation}
D_i(j;g)=\lVert z_{i\to j}-z_{i\to g}\rVert^2.
\end{equation}
A smaller $D_i(j;g)$ assigns higher priority to candidate $j$ under the learned goal-conditioned transition geometry.

\subsection{Bellman-Grounded Objective}
\label{sec:bellman_objective}

We train $\psi$ and $\phi$ using the Bellman-grounded regression objective
\begin{equation}
\mathcal{L}(\theta) = \mathbb{E}_{(i,j,g)}\Big[\big(D_i(j;g) - c_{i\to j} - \operatorname{sg}[\min_{k\in\mathcal{N}(j)}D_j(k;g)]\big)^2\Big]
\end{equation}
where $\operatorname{sg}[\cdot]$ denotes the stop-gradient operator. For nonterminal transitions, the target combines the immediate edge cost with the best downstream score, providing additive Bellman grounding. Detaching the bootstrap term yields a semi-gradient TD update \citep{sutton1988td}. The targets are constructed from sampled local transitions and observed edge costs, without shortest-path or optimal-action labels.

Because $D_i(g;g)=0$ by construction, the goal provides a structural terminal anchor for the recursion. In particular, when $g\in\mathcal{N}(j)$, the bootstrap term satisfies $\min_{k\in\mathcal{N}(j)}D_j(k;g)=D_j(g;g)=0$, so the target reduces to the immediate edge cost $c_{i\rightarrow j}$. Accordingly, we interpret $D_i(j;g)$ as a goal-conditioned transition-ranking score whose learned geometry reflects additive weighted route-cost structure, rather than as an exact optimal cost-to-go.

\subsection{Online Greedy Readout}
\label{sec:amortized_inference}
After training, BCM performs online planning through a local greedy rollout. The learned node representation is shared across goals, so changing the goal requires neither rebuilding nor retraining the map; it only changes the goal-conditioned scores used during local readout.

To reduce cycling during the rollout, at node $i$ we define the candidate set
\begin{equation}
\label{eq:candidate_set}
\mathcal{C}_i =
\begin{cases}
\mathcal{N}(i)\setminus\mathcal{V},
&
\mathcal{N}(i)\setminus\mathcal{V}\neq\varnothing,\\[2mm]
\mathcal{N}(i),
&
\text{otherwise},
\end{cases}
\end{equation}
where $\mathcal{V}$ contains the nodes visited during the current rollout. The next node is selected as
\begin{equation}
j^*
=
\arg\min_{j\in\mathcal{C}_i}
D_i(j;g).
\label{eq:readout}
\end{equation}

This visited-aware rule prefers unvisited neighbors while allowing revisits when no unvisited action remains. It introduces limited path memory without adding goal or shortest-path information; we isolate its effect in the ablation study.

\subsection{Compact Coordinate Encoding}
\label{sec:compact_encoding}

The node encoder $\psi$ maps node coordinates to embeddings. A naive state-wise one-hot code makes the input-layer parameter count grow as $\mathcal{O}(N)$, tying the encoder size to the number of nodes. For an $S\times S$ grid with $N=S^2$, we instead encode the two coordinate axes independently, reducing this parameter growth to $\mathcal{O}(S)=\mathcal{O}(\sqrt{N})$ for fixed embedding width. The resulting additive $k$-hot encoding, however, underperforms empirically (Table~\ref{tab:ablation-encoding}), suggesting that independent axis contributions do not adequately capture row--column interactions.

To introduce these interactions, let $r(n)=\lfloor n/S\rfloor$ and $c(n)=n\bmod S$ denote the row and column indices of node $n$. Learnable projections $W_r$ and $W_c$ map the corresponding one-hot coordinates to
\begin{equation}
\mathbf r_n = W_r \mathbf e_{r(n)},
\qquad
\mathbf c_n = W_c \mathbf e_{c(n)}.
\end{equation}
We then define
\begin{equation}
o_n
=
\psi(x_n)
=
\left[
\mathbf r_n;
\mathbf c_n;
\mathbf r_n\odot\mathbf c_n
\right]
\in\mathbb R^{d_o},
\end{equation}
where $\odot$ denotes the Hadamard product. The interaction term captures row--column combinations while preserving $\mathcal{O}(\sqrt{N})$ input-layer parameter growth. In the ablation study, this encoding substantially improves over additive $k$-hot encoding and approaches the performance of state-wise one-hot encoding.

\section{Experiments}

\subsection{Setup}
\label{sec:setup}

We evaluate on fixed weighted 2D grid graphs, where each node has integer coordinates and each edge has a known positive cost. We consider three obstacle layouts probing different geometric conditions: a U-shaped trap, a central block, and a vertical wall with a single-cell gap. The U-shaped layout scales proportionally with the grid side, whereas the wall thickness and gap in the wall-with-gap layout remain one cell wide. Edge costs follow three regimes: uniform (all costs equal to $1$), mild ($c_{ij}\sim\mathcal{U}(1,3)$), and wide ($c_{ij}\sim\mathcal{U}(1,10)$), with continuous uniform sampling in the latter two regimes. Unless otherwise stated, experiments use $N=1600$ ($40\times40$); scaling experiments vary $N\in\{400,900,1600,3600\}$ on the U-shaped layout under wide costs. 

Across graph sizes, BCM uses a transition MLP with two hidden layers of width $256$ and a $256$-dimensional output ($d_z=256$). The per-axis embedding dimension is $64$, yielding $d_o=192$ for the interaction encoding. Our main results (Table~\ref{tab:quality}) report mean$\pm$std over three runs using seeds $42$, $111$, and $512$, with $100$ evaluation start--goal pairs per run. Unless otherwise noted, other experiments use a single run and $100$ evaluation pairs per graph.

Our baselines span the space between searching anew for every goal and precomputing all answers in advance. At one end, Dijkstra provides exact per-query weighted-optimal search. At the other, APSP removes online search by precomputing exact pairwise distance and next-hop tables, requiring $\mathcal{O}(N^2)$ storage and enabling $\mathcal{O}(1)$ next-hop lookup at each rollout step. Between them lie methods that build a reusable representation once and read from it across goals. CML~\citep{stoeckl2024cml}, the direct predecessor of our method, learns an inverse-model metric giving heuristic directionality toward a goal. EigenAgent~\cite{zuo2026eigenagent} instead builds its representation analytically from the binary graph Laplacian, yielding a spectral potential that does not use edge costs. APF~\cite{khatib1986real} is a hand-designed local potential built from geometric attraction and obstacle repulsion, using only geometric distance rather than edge costs and requiring neither learning nor precomputation. Together, these methods differ along the three axes summarized in Table~\ref{tab:quality}.

Path quality is compared across Dijkstra, EigenAgent, CML, APF, and BCM. Because CML exhibits low success already on the base layout (Table~\ref{tab:quality}) and incurs a large state- and action-indexed representation (13.43M parameters at $N=1600$), we report it once as a lower reference and exclude it from the scale and layout analyses; the storage comparison covers BCM, EigenAgent, and APSP, representing parametric, spectral, and exact pairwise routing representations, respectively. BCM uses the interaction encoding throughout; its one-hot and $k$-hot variants appear in the ablation. For EigenAgent, we request $k=N-2$ modes at each size and retain all numerically converged modes; at $N=1600$, ARPACK converges to $1534$ modes. The main comparison uses the binary Laplacian, while two cost-weighted variants only partially reduce the empirical Gap. BCM uses visited-aware greedy readout in \eqref{eq:readout}, whereas CML, EigenAgent, and APF use their native greedy planners; Table~\ref{tab:ablation-visited} separately isolates the effect of BCM's visited-aware rule.

\paragraph{Metrics.}
We report success rate (SR), Cost, and Gap. SR is the fraction of start--goal pairs for which a method reaches the goal. Cost is the mean realized weighted path cost over successfully solved pairs. Gap is the mean per-pair relative excess over the corresponding Dijkstra optimum:
\begin{equation}
\mathrm{Gap}
=
\mathbb{E}\!\left[
\frac{c_{\mathrm{path}}}{c_{\mathrm{opt}}}-1
\;\middle|\;
\mathrm{success}
\right]\times100,
\label{eq:gap}
\end{equation}
where $c_{\mathrm{path}}$ is the realized path cost and $c_{\mathrm{opt}}$ is the Dijkstra-optimal cost for the same start--goal pair. Thus, $\mathrm{Gap}=0\%$ denotes optimal routing. Because Cost and Gap are conditioned on success, they must be interpreted jointly with SR when a method has incomplete success.

\subsection{Weighted Path Quality}
\label{sec:quality}
\begin{table*}[t]
\centering
\small
\setlength{\tabcolsep}{4pt}
\begin{tabular}{l ccc | ccc ccc ccc}
\toprule
& \multicolumn{3}{c|}{Properties}
& \multicolumn{3}{c}{Uniform}
& \multicolumn{3}{c}{Mild $[1,3]$}
& \multicolumn{3}{c}{Wide $[1,10]$} \\
\cmidrule(lr){2-4}
\cmidrule(lr){5-7}
\cmidrule(lr){8-10}
\cmidrule(lr){11-13}
Method
& Reuse & Weighted & Sublinear
& SR$\uparrow$ & Cost$\downarrow$ & Gap$\downarrow$
& SR$\uparrow$ & Cost$\downarrow$ & Gap$\downarrow$
& SR$\uparrow$ & Cost$\downarrow$ & Gap$\downarrow$ \\
\midrule
\textit{Dijkstra} \textit{(opt.)}
& \ding{55} & \ding{51} & \textemdash
& 1.00 & 28.5 & 0.0
& 1.00 & 47.1 & 0.0
& 1.00 & 108.8 & 0.0 \\
\midrule
CML$^{\dagger}$
& \ding{51} & \ding{51} & \ding{55}
& 0.27 & 40.0 & 185.6{\scriptsize $\pm$ 29.8}
& 0.20 & 50.6 & 147.8{\scriptsize $\pm$ 37.3}
& 0.16 & 107.1 & 154.4{\scriptsize $\pm$ 72.0} \\
APF$^{\dagger}$
& \ding{55} & \ding{55} & \textemdash
& 0.82 & 27.0 & 0.0{\scriptsize $\pm$ 0.0}
& 0.82 & 53.8 & 19.7{\scriptsize $\pm$ 0.3}
& 0.82 & 147.4 & 42.3{\scriptsize $\pm$ 0.8} \\
EigenAgent
& \ding{51} & \ding{55} & \ding{55}$^{\ast}$
& 1.00 & 29.9 & 4.4{\scriptsize $\pm$ 1.1}
& 1.00 & 59.5 & 24.4{\scriptsize $\pm$ 1.7}
& 1.00 & 163.0 & 47.2{\scriptsize $\pm$ 1.7} \\
BCM (ours)
& \ding{51} & \ding{51} & \ding{51}
& 1.00 & \textbf{28.5} & \textbf{0.07}{\scriptsize $\pm$ 0.05}
& 1.00 & \textbf{48.8} & \textbf{4.01}{\scriptsize $\pm$ 0.37}
& 1.00 & \textbf{114.0} & \textbf{5.07}{\scriptsize $\pm$ 0.64} \\
\bottomrule
\end{tabular}
\caption{
Path quality under increasing edge-cost heterogeneity ($N=1600$, U-shaped layout). \emph{Reuse} denotes reuse across goals; \emph{Weighted}, the use of edge costs; and \emph{Sublinear}, reusable representation storage growing sublinearly in $N$ (\textemdash: not applicable). $^{\ast}$ EigenAgent requests $k=N-2$ modes and retains $1534$ numerically converged modes at $N=1600$, yielding quadratic storage in this near-full-rank setting. Results are mean$\pm$std over three runs, with $100$ pairs per run. Cost and Gap are computed over successful pairs; $^{\dagger}$ marks incomplete success. Best full-success non-reference results are shown in \textbf{bold}. CML uses the authors' released implementation and native greedy planner.
}
\label{tab:quality}
\end{table*}

We first test whether grounding a reusable cognitive map in additive edge costs improves routing as cost heterogeneity increases. Table~\ref{tab:quality} compares the methods on the same U-shaped graph under uniform, mild, and wide costs. Under uniform costs, EigenAgent and BCM both reach every goal with low Gap. As costs become heterogeneous, the Gap of the binary-Laplacian EigenAgent increases from $24.4\%$ under mild costs to $47.2\%$ under wide costs, whereas BCM remains at $4.0\%$ and $5.1\%$, respectively. APF fails on $18\%$ of the pairs across all three regimes and reaches a $42.3\%$ Gap under wide costs on the pairs it solves. CML exhibits low success across all three regimes, despite receiving edge-cost information in the evaluated weighted setting.

These results highlight different sensitivities among the evaluated alternatives. APF's local geometric potential is susceptible to the U-shaped trap and does not account for heterogeneous edge costs. Although the evaluated weighted CML variant receives edge-cost information, its directional objective does not explicitly encode additive downstream route cost. EigenAgent retains full success, but its binary-Laplacian representation encodes connectivity without using edge costs, consistent with its increasing Gap as cost heterogeneity grows. BCM instead grounds its transition scores in immediate edge costs and bootstrapped downstream scores, maintaining low empirical Gap across the tested weight regimes.

One remaining possibility is that EigenAgent's degradation arises simply because the main comparison uses a binary Laplacian. We therefore evaluate two cost-weighted variants: inverse-cost and RBF-weighted Laplacians. Under mild costs, their Gap decreases only from $24.4\%$ to $21.4\%$--$22.9\%$. Under wide costs, the inverse-cost variant reaches $42.2\%$, while the RBF variant reaches $51.8\%$. Neither tested variant closes the gap to BCM. For the evaluated constructions, these results suggest that incorporating costs into spectral affinities alone is insufficient to reproduce the low-Gap behavior obtained through BCM's additive downstream grounding.

\paragraph{Performance across layouts.}
We next test whether the effect of cost grounding persists across different obstacle geometries. We repeat the comparison on a central-block layout and a wall with a single-cell gap under mild and wide costs (Table~\ref{tab:generalization}). BCM reaches every goal on all three layouts, with Gap remaining below $5.9\%$. The binary-Laplacian EigenAgent again degrades as cost heterogeneity increases, reaching $44.6\%$--$47.0\%$ Gap under wide costs. APF is more sensitive to obstacle geometry: its success rate ranges from $0.98$ on the central-block layout to $0.68$ on the wall-with-gap layout, while its Gap also increases with cost heterogeneity. Across the tested layouts, BCM therefore maintains low Gap and full success relative to the evaluated geometric and connectivity-based alternatives.

\begin{table}[t]
\centering
\small
\setlength{\tabcolsep}{1.0pt} 
\renewcommand{\arraystretch}{0.95}
\begin{tabular*}{\linewidth}{@{\extracolsep{\fill}} l l ccc ccc @{}}
\toprule
& & \multicolumn{3}{c}{Mild $[1,3]$} & \multicolumn{3}{c}{Wide $[1,10]$} \\
\cmidrule(lr){3-5}\cmidrule(lr){6-8}
Layout & Method & SR$\uparrow$ & Cost$\downarrow$ & Gap$\downarrow$ & SR$\uparrow$ & Cost$\downarrow$ & Gap$\downarrow$ \\
\midrule
\multirow{4}{*}{U-shaped$^{\ast}$}
  & Dijkstra \textit{(opt.)} & 1.00 & 47.3 & 0.0 & 1.00 & 106.4 & 0.0 \\
  & APF$^{\dagger}$ & 0.78 & 51.6 & 20.1 & 0.78 & 130.4 & 42.2 \\
  & EigenAgent & 1.00 & 58.5 & 23.6 & 1.00 & 158.3 & 45.2 \\
  & BCM & \textbf{1.00} & \textbf{48.9} & \textbf{3.5} & \textbf{1.00} & \textbf{111.3} & \textbf{5.8} \\
\midrule
\multirow{4}{*}{Center block}
  & Dijkstra \textit{(opt.)} & 1.00 & 42.9 & 0.0 & 1.00 & 97.9 & 0.0 \\
  & APF$^{\dagger}$ & 0.98 & 53.4 & 22.7 & 0.98 & 146.5 & 47.2 \\
  & EigenAgent & 1.00 & 53.2 & 21.8 & 1.00 & 144.9 & 44.6 \\
  & BCM & \textbf{1.00} & \textbf{44.2} & \textbf{3.2} & \textbf{1.00} & \textbf{102.5} & \textbf{5.2} \\
\midrule
\multirow{4}{*}{Wall w/ gap}
  & Dijkstra \textit{(opt.)} & 1.00 & 48.1 & 0.0 & 1.00 & 110.4 & 0.0 \\
  & APF$^{\dagger}$ & 0.68 & 46.3 & 21.1 & 0.68 & 126.4 & 43.9 \\
  & EigenAgent & 1.00 & 60.9 & 24.4 & 1.00 & 165.7 & 47.0 \\
  & BCM & \textbf{1.00} & \textbf{49.5} & \textbf{3.0} & \textbf{1.00} & \textbf{114.7} & \textbf{5.3} \\
\bottomrule
\end{tabular*}
\caption{
Path quality across obstacle layouts under mild and wide edge-cost heterogeneity ($N=1600$). Cost and Gap are computed over successfully solved pairs. $^{\dagger}$ denotes methods with $\mathrm{SR}<1$. $^{\ast}$ The U-shaped rows use a single seed ($42$), whereas Table~\ref{tab:quality} reports three-seed means. Bold indicates the lowest Cost and Gap among non-reference methods with full success.
}
\label{tab:generalization}
\end{table}

\subsection{Scalability and Storage}
\label{sec:scale}

We next examine whether BCM's compact cost-grounded representation maintains low empirical Gap as graph size increases. We vary $N\in\{400,900,1600,3600\}$ on the U-shaped layout under wide costs, using the same architecture and hyperparameters throughout.

\begin{figure*}[t]
    \centering
    \includegraphics[width=\textwidth]{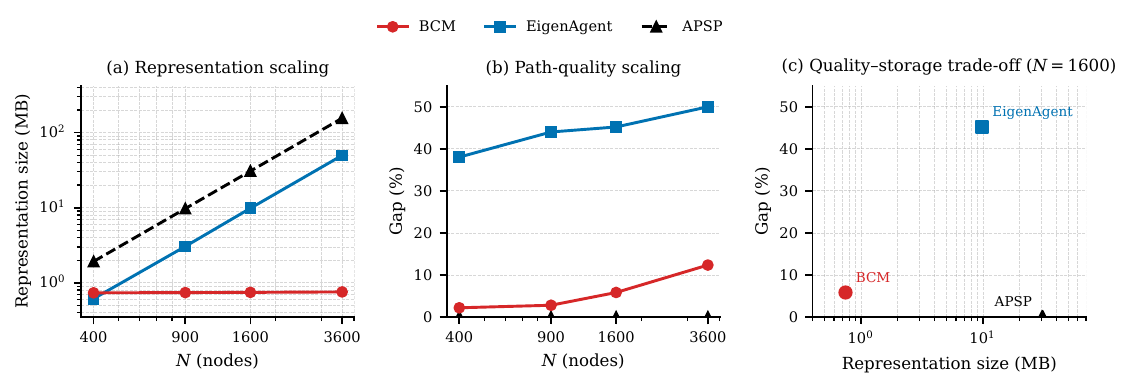}
    \caption{Scalability and method-specific representation size on the U-shaped layout under wide edge costs. (a) Representation size on a logarithmic scale, counting model parameters for BCM, retained eigenvectors and eigenvalues for EigenAgent, and exact distance and next-hop tables for APSP. Optional BCM node-embedding caches, graph adjacency, and edge-cost storage are excluded. (b) Path-quality Gap as $N$ increases. (c) Empirical quality--storage trade-off at $N=1600$. All plotted methods reach every goal at every tested size; APSP is the exact quadratic-storage reference.}
    \label{fig:quantitative}
\end{figure*}

\paragraph{Storage.}
Figure~\ref{fig:quantitative}(a) shows that BCM's representation grows only from $0.734$~MB at $N=400$ to $0.755$~MB at $N=3600$. Its transition encoder contributes a fixed $0.724$~MB, while the coordinate-dependent parameters grow as $\mathcal{O}(\sqrt{N})$. In contrast, EigenAgent's near-full-rank spectral representation grows from $0.613$~MB to $49.751$~MB, and the exact distance and next-hop tables of APSP grow from $1.920$~MB to $155.520$~MB. At $N=1600$, EigenAgent and APSP require $13.2\times$ and $41.3\times$ the representation size of BCM, respectively; at $N=3600$, these ratios increase to $65.9\times$ and $206.1\times$.

\paragraph{Path quality and trade-off.}
BCM maintains full success at every tested size, with Gap remaining below $5.9\%$ through $N=1600$ and rising to $12.4\%$ at $N=3600$. EigenAgent also retains full success, but its Gap remains above $38\%$ and reaches $50.0\%$ at the largest size (Figure~\ref{fig:quantitative}(b)). At $N=1600$, BCM achieves $5.8\%$ Gap with a $0.744$~MB representation, compared with $45.2\%$ Gap and $9.824$~MB for EigenAgent; APSP attains exact routing with $30.720$~MB (Figure~\ref{fig:quantitative}(c)). Among the evaluated reusable representations, BCM therefore provides a substantially lower-Gap and smaller-representation operating point through $N=1600$. Its degradation at $N=3600$, however, indicates reduced accuracy at the largest tested scale.

\subsection{Ablation}
\label{sec:ablation}
We isolate two design choices that support BCM's empirical performance: the interaction-based coordinate encoding and the visited-aware greedy readout. All experiments use the U-shaped layout with $N=1600$.

\paragraph{Coordinate encoding.}
Table~\ref{tab:ablation-encoding} isolates the node encoding while holding the transition head, training protocol, and readout fixed. The state-wise one-hot encoding attains the lowest Gap, but its coordinate-dependent parameter count grows as $\mathcal{O}(N)$, yielding $541$K total parameters at $N=1600$. The additive $k$-hot encoding reduces the model to $152$K parameters but degrades substantially under heterogeneous costs. Adding the cross-axis interaction increases the parameter count by $22\%$, from $152$K to $186$K, while reducing Gap from $12.82\%$ to $3.31\%$ under mild costs and from $15.88\%$ to $4.94\%$ under wide costs. The interaction encoding therefore approaches one-hot performance within $0.9$ percentage points while using about one third of its parameters and retaining $\mathcal{O}(\sqrt{N})$ coordinate-dependent parameter growth.

\begin{table}[t]
\centering
\small
\setlength{\tabcolsep}{2.5pt}
\renewcommand{\arraystretch}{1.05}
\begin{tabular}{@{} l c r cc @{}}
\toprule
Encoding & Scaling & \# Params & Gap$\downarrow$ (mild) & Gap$\downarrow$ (wide) \\
\midrule
one-hot & $\mathcal{O}(N)$ & 541K & \textbf{2.55} & \textbf{4.11} \\
$k$-hot (additive) & $\mathcal{O}(\sqrt{N})$ & 152K & 12.82 & 15.88 \\
interaction (ours) & $\mathcal{O}(\sqrt{N})$ & 186K & 3.31 & 4.94 \\
\bottomrule
\end{tabular}
\caption{
Coordinate-encoding ablation ($N=1600$, U-shaped layout; Gap in \%). All variants share the same transition head, training protocol, and readout; only the node encoding changes. All variants use a single seed (42), whereas Table~\ref{tab:quality} reports three-seed means. \emph{Scaling} denotes asymptotic parameter growth with $N$, whereas \# Params gives the total trainable parameter count at $N=1600$. Bold indicates the lowest Gap.
}
\label{tab:ablation-encoding}
\end{table}
\begin{table}[t]
\centering
\small
\setlength{\tabcolsep}{4pt}
\begin{tabular}{l cc cc}
\toprule
& \multicolumn{2}{c}{SR$\uparrow$} & \multicolumn{2}{c}{wSPL$\uparrow$} \\
\cmidrule(lr){2-3}\cmidrule(lr){4-5}
Encoding & None & Visited-aware & None & Visited-aware \\ 
\midrule
one-hot           & 0.86 & 1.00 & 0.833 & 0.958 \\
$k$-hot            & 0.76 & 0.97 & 0.698 & 0.865 \\
interaction (ours) & 0.84 & 1.00 & 0.799 & 0.950 \\
\bottomrule
\end{tabular}
\caption{Visited-aware readout ablation on the U-shaped layout under the
wide-cost regime ($N=1600$). We report weighted SPL,
$\mathrm{wSPL}=\frac{1}{M}\sum_{i=1}^{M}
S_i\,\frac{c_i^{\mathrm{opt}}}
{\max(c_i^{\mathrm{opt}},c_i^{\mathrm{path}})}$,
where $M$ is the number of Dijkstra-reachable evaluation queries,
$S_i\in\{0,1\}$ is the success indicator, and
$c_i^{\mathrm{opt}}$ and $c_i^{\mathrm{path}}$ are the Dijkstra-optimal
and realized weighted path costs, respectively.
Failed rollouts contribute zero.
This metric adapts SPL~\citep{anderson2018spl} by replacing path length
with weighted cost. Gap is omitted because it is conditioned on success
and is not directly comparable when SR differs substantially.}
\label{tab:ablation-visited}
\end{table}

\paragraph{Visited-aware readout.} Without path memory, greedy local readout can repeatedly select previously visited nodes and enter short cycles. The visited-aware rule prefers unvisited neighbors and falls back to the full neighborhood when no unvisited candidate remains. Applying this rule without changing the learned parameters raises the interaction model's SR from $0.84$ to $1.00$ under wide costs. The improvement holds across all three encodings, ranging from $14$ to $21$ percentage points. At the same time, wSPL (defined in Table~\ref{tab:ablation-visited}) rises from $0.799$ to $0.950$, while the average weighted-cost efficiency among successful rollouts remains approximately $0.95$. Thus, the improvement in success does not come at the cost of more expensive successful routes. The rule adds no learned parameters and requires only a temporary length-$N$ visited mask during each rollout. The ablation therefore shows that the learned cost-grounded scores can support efficient routes, while lightweight path memory improves the robustness of greedy local readout by reducing short cycles.

\subsection{Limitations} \label{sec:limitations} 
BCM currently targets repeated planning on fixed, known, and coordinate-structured weighted graphs, with one representation learned per graph. Substantial changes to topology or edge costs would therefore require updating or retraining the model. We also observe reduced accuracy on denser maze layouts and at the largest tested scale: under wide costs, Gap rises to $12.4\%$ at $N=3600$. These results indicate that the current model is most reliable when coordinates provide meaningful structural information and at moderate graph scales. 

As a learned local planner, BCM does not inherit the formal optimality or completeness guarantees of exact graph search. In addition, the current terminal convention supports goal-conditioned transition ranking rather than calibrated final-edge value estimation.

\section{Related Work}

\paragraph{Cognitive maps.}
Cognitive maps~\cite{tolman1948cognitive} are internal representations of an environment that support flexible routing, studied both as accounts of biological navigation~\cite{whittington2020tem,gornet2024predictive} and as reusable structures for planning. Existing approaches differ in the quantity encoded by the representation. Successor representations~\cite{dayan1993sr,stachenfeld2017sr,barreto2017sf} learn policy-conditioned future occupancy through an expectation-based temporal-difference recursion. Spectral methods~\citep{mahadevan2007pvf,machado2018eigenoption,wu2019laplacian} organize states through diffusion in the graph-Laplacian eigenbasis, including EigenAgent~\citep{zuo2026eigenagent}. Our main EigenAgent instantiation is constructed from the binary Laplacian and therefore does not use heterogeneous edge costs; the two cost-weighted variants only partially reduce the empirical Gap. Inverse-model metrics~\cite{stoeckl2024cml,lin2026gcml,polykretis2024mapless} learn embedded observation differences aligned with corresponding action vectors, producing an action-aligned directional signal for local guidance without an explicit additive downstream-cost recursion. Linear RL~\cite{piray2021linearrl} and its featurized extensions~\cite{bazarjani2026dfr} are closer in their use of a Bellman formulation for control, but solve a default-policy-regularized soft-control problem through a linearized formulation. Across these lines, BCM differs by using immediate edge costs and a hard minimum over downstream transitions, grounding its nonterminal transition-ranking scores in a hard-minimum Bellman-grounded recursion over the graph's edge costs.

\paragraph{Learned planning representations.}
Goal-conditioned reinforcement learning~\cite{kaelbling1993goals}, including universal value functions $V(s,g)$~\cite{schaul2015uvfa}, reuses learned parameters across changing goals. Quasimetric representations similarly support goal reaching through greedy descent~\cite{wang2023qrl,eysenbach2022contrastive}. Differentiable planners~\cite{tamar2016vin,wang2024highwayvin,wang2025dtvin,zhao2023implicit} embed value-iteration-like computation in learned architectures, while Plan2vec~\cite{yang2020plan2vec} learns a latent representation for planning using shortest-path supervision. BCM focuses on a different operating point: repeated planning on one fixed, known weighted graph. It learns a compact, graph-specific, cost-grounded representation from local graph transitions and edge costs, without shortest-path labels or per-goal retraining.

\paragraph{Search.}
Dijkstra and A\textsuperscript{*} with an admissible heuristic~\cite{dijkstra1959,hart1968astar} provide exact per-query planning. All-pairs precomputation~\cite{floyd1962apsp} shifts computation into a quadratic-storage table. Contraction hierarchies~\cite{geisberger2008ch} trade graph preprocessing and auxiliary storage for faster exact shortest-path queries. Learned heuristics~\cite{yonetani2021neurala,archetti2021nwastar} retain per-query search while reducing node expansions, generally without retaining the same admissibility-based guarantees. BCM studies a different operating point: a graph-specific learned representation is constructed once and reused across goals through local readout. We therefore use exact search and all-pairs precomputation as references; unlike exact graph search, BCM does not provide formal optimality or completeness guarantees.

\section{Discussion}
BCM separates planning into a few parts—a coordinate encoding, a transition encoder, and a local readout—all trained under a single Bellman-grounded objective. The objective provides the common training principle; the surrounding components are implementation choices, and here each is instantiated in a basic form: a compact coordinate encoding whose parameters grow with per-axis resolution, a feedforward transition encoder with two hidden layers, and a greedy readout with a visited-aware rule. That so minimal an instance already achieves low empirical Gap suggests that cost-grounded representation is a useful core ingredient, while each surrounding component remains a place to strengthen rather than a fixed part of the method.

Consider the encoding. Its binding principle---encoding factors separately and then binding them to distinguish their conjunctions---is not conceptually specific to coordinates and may extend to other settings where node identity factorizes: grid coordinates split naturally into $x$ and $y$, while a puzzle state may decompose into tile identity and position. Coordinates are the case studied here, and our experiments validate this principle only for 2D grids; nevertheless, extending such factors toward more abstract structure offers a promising direction toward more transferable representations. The payoff of this binding is structural: the transition encoder remains fixed in width, the readout adds no learned parameters, and the coordinate-dependent model parameters grow as $\mathcal{O}(\sqrt{N})$. Consequently, BCM's method-specific representation increases only from $0.734$~MB at $N=400$ to $0.755$~MB at $N=3600$, while supporting goal-conditioned transition scores that reflect weighted route-cost structure. Beyond the coordinate encoding, the transition encoder is equally open to strengthening: a graph-based encoder aggregating neighborhood structure could provide topological cues that the current feedforward map must infer indirectly.

The readout admits a similar interpretation. The greedy rule follows the network's goal-conditioned transition scores without memory of its path and can stall by cycling among a few nodes. A visited-aware rule that prioritizes unvisited neighbors, while allowing fallback when none remain, substantially improves success without retraining or changing the learned scores. That such a lightweight modification is effective suggests that memoryless rollout contributes substantially to the remaining failures. Stronger readouts incorporating limited lookahead, backtracking, or local search are therefore a natural next step, while the reusable cost-grounded representation remains the common foundation.

Taken together, these components make BCM a minimal but complete instance: strong enough to achieve low empirical Gap on the tested weighted graphs, plain enough that each part invites a better replacement. We offer it less as a finished planner than as a modular starting point for representation-centric planning.

\section{Conclusion}
\label{sec:conclusion}

We presented BCM, a cost-grounded cognitive-map model for repeated planning
on fixed, known weighted graphs. BCM learns goal-conditioned
transition-ranking scores from local heterogeneous edge costs through a
Bellman-grounded objective and generates routes using visited-aware greedy
readout. Across the tested weighted grids, BCM maintains full success and
approximately $5\%$ mean Gap through $N=1600$ under the widest weight range,
compared with about $45\%$ for a representative connectivity-based spectral
map. Its coordinate-dependent parameter count grows as
$\mathcal{O}(\sqrt{N})$ on two-dimensional grids, although Gap rises to
about $12\%$ at $N=3600$. Because the underlying representation is shared
across goals, changing the queried goal requires neither rebuilding the map nor per-goal
retraining. These results show that additive route-cost structure can be
encoded in a compact, reusable cognitive-map representation.

\section*{Ethical Statement}
This work studies graph-planning algorithms in synthetic fixed environments
and does not use human-subject data, private data, or deployed decision
systems. Potential risks are limited to downstream use of planning systems
in real-world settings, where safety constraints, uncertainty, and failure
recovery should be evaluated separately before deployment.

\bibliography{BCM}

\end{document}